%% file: main.tex
\documentclass[11pt,a4paper]{article}
\usepackage[a4paper,margin=1in]{geometry}
\usepackage[T1]{fontenc}
\usepackage[utf8]{inputenc}
\usepackage{newtxtext,newtxmath}
\usepackage{graphicx}
\usepackage{float}
\usepackage{tikz}
\usetikzlibrary{arrows.meta,positioning,calc}
\usepackage{fontawesome5}
\usepackage{booktabs}
\usepackage{tabularx}
\usepackage{array}
\usepackage{enumitem}
\usepackage{microtype}
\usepackage{xurl}
\usepackage[hidelinks]{hyperref}
\usepackage{textcomp}
\graphicspath{{figures/}}
\setlist{nosep,leftmargin=*}
\hypersetup{
  pdftitle={Harness Robotic OS: A Unified Embodied-Agent Runtime for Closed-Loop Quadruped Inspection},
  pdfauthor={Yaoyuan Yan, Zhiyou Heng, Haoxiang Jie, Gang Liu, Hongjie Yan, Wei Zhou}
}

\newcommand{\makepaperheading}{%
  \begingroup
  \centering
  \vspace*{-0.7cm}
  \noindent\rule{\textwidth}{1.15pt}\par
  \vspace{0.75em}
  \begin{minipage}{0.96\textwidth}
    \centering
    {\LARGE\bfseries\textit{Harness Robotic OS}: A Unified Embodied-Agent Runtime\\[0.18em]
     for Closed-Loop Quadruped Inspection\par}
  \end{minipage}
  \vspace{0.75em}
  \noindent\rule{\textwidth}{1.15pt}\par
  \vspace{1.25em}
  {\large\scshape A Preprint\par}
  \vspace{1.35em}
  {\large\bfseries
    Yaoyuan Yan\textsuperscript{1}\quad
    Zhiyou Heng\textsuperscript{1}\quad
    Haoxiang Jie\textsuperscript{1}\par
    \vspace{0.22em}
    Gang Liu\textsuperscript{1}\quad
    Hongjie Yan\textsuperscript{2,3}\quad
    Wei Zhou\textsuperscript{1}\par}
  \vspace{0.75em}
  {\normalsize
    \textsuperscript{1}AI Lab, Country Garden Services\quad
    \textsuperscript{2}Omni AI\quad
    \textsuperscript{3}East China Normal University\par}
  \vspace{0.35em}
  {\small\ttfamily
    \{yanyaoyuan,hengzhiyou,jiehaoxiang\}@bgyfw.com\quad
    51280147091@stu.ecnu.edu.cn\par}
  \vspace{1.15em}
  {\normalsize \today\par}
  \endgroup
}

\renewenvironment{abstract}{%
  \vspace{1.0em}
  \begin{center}{\large\bfseries Abstract}\end{center}
  \vspace{-0.35em}
  \begin{quote}\noindent
}{%
  \end{quote}
}

\begin{document}
\makepaperheading

\begin{abstract}
Autonomous property inspection requires more than robust robot navigation: a deployable system must connect heterogeneous sensing, reusable autonomy capabilities, multimodal scene understanding, human interaction, and enterprise response within a traceable operational loop. Existing quadruped inspection systems commonly integrate these functions through task-specific interfaces, making contextual coordination, knowledge reuse, and controlled adaptation difficult. This paper presents \textit{Harness Robotic OS} (HROS), a unified embodied-agent runtime, and Argos, its realization for residential-community inspection. HROS organizes the system into robot runtime, embodied autonomy skills, cognitive agent runtime, and interaction and operations planes. A shared context connects physical state with agent reasoning; streaming ASR/TTS supports voice-based mission interaction; hierarchical working, episodic, and semantic memory preserves operational knowledge; and a safety-gated self-evolution loop converts execution traces into versioned candidate updates without permitting unconstrained online modification. The Argos prototype integrates a Vbot quadruped, Fast-LIO2 localization and mapping, Hobot-Stereo depth perception, PCT-Planner global planning, EGO-Planner local motion generation, and OpenClaw-orchestrated Qwen3-VL inspection analysis. Experiments in a residential property environment achieved 100\% waypoint reachability, outdoor localization error below 10~cm, local obstacle-response latency below 200~ms, representative hazard-detection rates of 85--95\%, and 99\% success in alarm delivery and structured-report generation. These results validate the deployed navigation and inspection closed loop, while HROS provides an extensible software foundation for memory-augmented, voice-aware, and continuously improvable embodied inspection agents.
\end{abstract}


\section{Introduction}

Residential property inspection spans roads, fire passages, building entrances, equipment rooms, waste-collection areas, and other public spaces. These environments combine large spatial coverage with changing illumination, pedestrians and vehicles, temporary obstacles, and diverse safety and sanitation risks. Manual patrols remain effective at contextual judgment, but their frequency, consistency, and traceability are constrained by labor and individual experience. A useful autonomous system must therefore provide repeatable physical coverage and convert observations into actionable, auditable property-management outcomes.

Quadruped robots are a promising carrier for this task. Compared with wheeled platforms, which are most effective on regular surfaces, legged systems can better negotiate ramps, thresholds, uneven pavement, standing water, and temporarily cluttered passages \cite{ref1,ref2}. Platforms such as ANYmal and Boston Dynamics Spot have demonstrated the mobility and payload capacity required for autonomous inspection in challenging environments \cite{ref3,ref4,ref5}. However, locomotion alone does not constitute an inspection system. Reliable deployment also requires persistent localization, global mission planning, reactive collision avoidance, scene-level hazard understanding, operator interaction, and integration with downstream work-order processes.

Most practical implementations treat these capabilities as a collection of loosely coupled modules. Sensor drivers, navigation algorithms, vision-language services, operator interfaces, and enterprise applications often maintain separate state and communicate through task-specific adapters. This fragmentation creates three limitations. First, semantic task intent is easily disconnected from robot pose, observations, and execution state. Second, experience from previous missions is not systematically retained and retrieved, limiting long-term adaptation. Third, changes to prompts, tool policies, task graphs, or reusable skills are difficult to evaluate, trace, and roll back safely. These limitations become especially important when an inspection robot must operate repeatedly in the same property while collaborating with both on-site personnel and remote operators.

We address this system-level gap through \textit{Harness Robotic OS} (HROS), an embodied-agent runtime that treats sensing, autonomy, reasoning, interaction, and operational feedback as composable capabilities within one governed loop. HROS maintains shared context across the physical and cognitive layers, exposes navigation and perception functions as reusable agent skills, and supports multimodal task orchestration through OpenClaw. Voice input is grounded through ASR against the current mission and robot state, while TTS returns confirmations, progress, alerts, and recovery information. A hierarchical memory separates short-horizon working context, time-indexed mission episodes, and stable semantic site knowledge. Rather than allowing uncontrolled online model changes, HROS uses an experience-to-update pipeline in which candidate memory, prompt, tool, or task-graph adaptations must pass offline evaluation and a versioned safety gate.

\begin{figure}[t]
\centering
\resizebox{\textwidth}{!}{\input{figures/system_architecture.tex}}
\caption{Harness Robotic OS architecture. The cognitive runtime combines voice and multimodal interaction, hierarchical agent memory, reusable embodied skills, and a safety-gated self-evolution loop that transforms execution experience into validated runtime updates.}
\label{fig:architecture}
\end{figure}
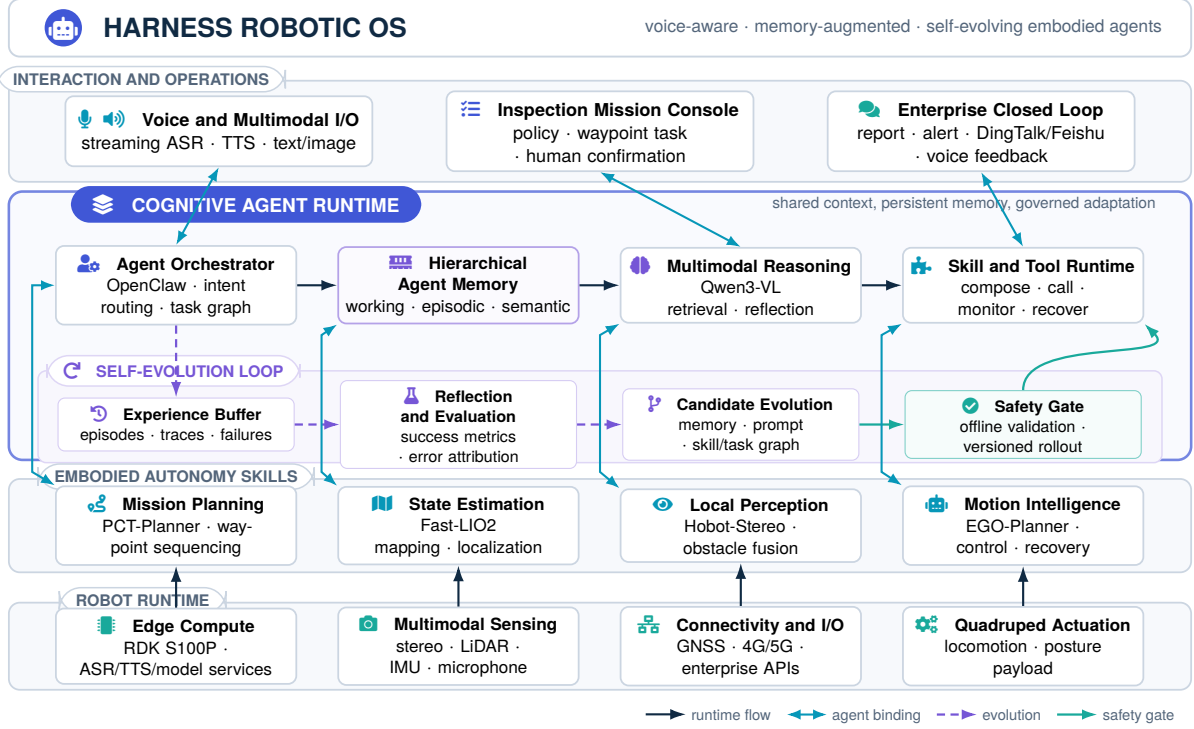

Argos instantiates HROS on a Vbot quadruped for residential-community inspection. Fast-LIO2 provides LiDAR--inertial mapping and localization; Hobot-Stereo complements LiDAR with near-field depth; PCT-Planner generates global inspection routes; and EGO-Planner performs local trajectory optimization and obstacle avoidance. At designated viewpoints, OpenClaw packages images, pose, and task context for Qwen3-VL, then associates returned hazard semantics with time and location before generating structured reports and enterprise alerts. The overall architecture is summarized in Fig.~\ref{fig:architecture}. This implementation enables the paper to evaluate the physical navigation and inspection closed loop while defining a scalable architecture for voice-aware, memory-augmented, and governed self-evolving agents.

The main contributions of this work are summarized as follows:

\begin{enumerate}
\item We propose HROS, a layered embodied-agent runtime that unifies robot resources, autonomy skills, shared context, multimodal reasoning, and property-management operations in a traceable closed-loop architecture.
\item We introduce a cognitive runtime that combines grounded ASR/TTS interaction, hierarchical working--episodic--semantic memory, and a safety-gated self-evolution process with evaluation, provenance, versioned rollout, and rollback.
\item We realize Argos by integrating Fast-LIO2 \cite{ref9}, Hobot-Stereo \cite{ref11}, PCT-Planner \cite{ref10}, and EGO-Planner \cite{ref12} on a Vbot quadruped, providing a complete mapping, localization, global planning, local perception, and motion pipeline for residential environments.
\item We implement an OpenClaw-orchestrated multimodal inspection workflow \cite{ref13} that connects Qwen-based hazard understanding \cite{ref14} with time--location grounding, structured reporting, and enterprise alerts, and evaluate the resulting end-to-end system in a real residential property environment.
\end{enumerate}

\section{Related Work}

The proposed system lies at the intersection of legged-robot autonomy, three-dimensional planning, multimodal inspection, and embodied-agent orchestration. We review these areas with emphasis on the integration gap addressed by HROS.

\subsection{Localization and Mapping}

Recent studies in localization and mapping mainly focus on LiDAR-based SLAM, vision-based SLAM, and multi-sensor fusion SLAM approaches. Representative LiDAR-based SLAM algorithms include Gmapping \cite{ref15} and Hector-SLAM \cite{ref16}, which have been widely applied and evaluated in indoor 2D environments. Representative vision-based SLAM algorithms include ORB-SLAM \cite{ref17}, ORB-SLAM2 \cite{ref18}, and VINS-Mono \cite{ref19}. In recent years, LOAM \cite{ref20}, LIO-SAM \cite{ref21}, and FAST-LIO \cite{ref22} further improved LiDAR-inertial fusion performance in terms of accuracy and robustness. Among recent representative methods, Fast-LIO2 is a typical LiDAR-Inertial Odometry (LIO) algorithm that fuses LiDAR and IMU information to achieve high-precision state estimation with relatively low computational overhead while effectively reducing cumulative errors caused by single sensors. Therefore, this paper adopts the Fast-LIO2 algorithm for localization and mapping.

\subsection{Global Path Planning}

In recent years, researchers have proposed various classical and improved algorithms for global path planning, focusing on planning efficiency, path smoothness, and adaptability to complex three-dimensional environments. Among traditional graph-search methods, the Dijkstra algorithm \cite{ref23} provides good optimality but suffers from relatively low computational efficiency. The A* algorithm \cite{ref24} significantly improves search efficiency by introducing heuristic functions and has become one of the most widely used global planning algorithms in mobile robotics. The D* algorithm \cite{ref25} further enhances replanning capability in dynamic environments. In addition, Theta* \cite{ref26} and Jump Point Search (JPS) \cite{ref27} improve A* in terms of path smoothness and search speed, respectively.

As robotic systems continue to increase in motion freedom, sampling-based algorithms such as RRT \cite{ref28}, RRT* \cite{ref29}, PRM \cite{ref30}, and BIT* \cite{ref31} can rapidly generate feasible paths in high-dimensional spaces; however, the generated trajectories are often unsmooth and exhibit slow convergence. Subsequently, methods such as Hybrid A* \cite{ref32}, Kinodynamic A* \cite{ref33}, and State Lattice \cite{ref34} began incorporating robotic kinematic and dynamic constraints into the path-searching process. More recently, for complex three-dimensional environments, methods such as MPC Planner \cite{ref35}, CHOMP \cite{ref36}, STOMP \cite{ref37}, TrajOpt \cite{ref38}, and PCT-Planner have achieved higher-quality trajectory generation through continuous-time trajectory modeling and optimization. PCT-Planner not only provides high path smoothness but also effectively incorporates dynamic constraints. Moreover, it demonstrates strong adaptability to 3D point cloud maps. Therefore, this paper adopts PCT-Planner for global path planning.

\subsection{Local Path Planning}

In recent years, local path planning has gradually evolved from traditional reactive obstacle avoidance methods toward approaches that consider dynamic constraints, continuous trajectory optimization, and dynamic environment prediction. Among traditional methods, the Dynamic Window Approach (DWA) \cite{ref39} has been widely applied in ROS navigation systems due to its simplicity and high real-time performance; however, it is prone to generating unsmooth trajectories and local oscillations in complex scenarios. Artificial Potential Field (APF) methods \cite{ref40} feature low computational complexity and simple implementation but easily fall into local optima.

With increasing environmental complexity and robot performance requirements, the Timed Elastic Band (TEB) method \cite{ref41} performs trajectory optimization through a time-parameterized elastic band model, balancing path smoothness and motion constraints, though it strongly depends on parameter tuning and global path quality. Model Predictive Control (MPC) \cite{ref42} achieves predictive control through receding-horizon optimization but imposes high computational demands. In addition, MPPI \cite{ref43}, reinforcement learning (RL)-based local planning \cite{ref44}, and learning-based prediction methods combined with MPC \cite{ref45} have gradually become research hotspots.

EGO-Planner avoids the explicit construction of traditional ESDF maps and directly utilizes local obstacle distance information for trajectory optimization, thereby significantly reducing computational overhead while maintaining trajectory smoothness and dynamic feasibility. Owing to its strong real-time performance and robustness, this paper adopts the EGO-Planner algorithm for local path planning.

\subsection{Multimodal Inspection and Embodied Agents}

Vision-language models extend robotic perception from predefined object categories to open-vocabulary scene interpretation and instruction-conditioned reasoning \cite{ref6,ref7,ref8}. This capability is relevant to property inspection, where hazards are often defined by spatial and operational context rather than object identity alone. A blocked fire passage, for example, requires jointly interpreting the observed objects, their location, and the applicable inspection rule. However, a remote multimodal model does not by itself provide mission control, robot-state grounding, persistent memory, or reliable enterprise integration.

Agent frameworks can coordinate model calls and external tools, but directly applying a generic agent loop to a mobile robot leaves unresolved questions of timing, safety, state consistency, and recovery. HROS therefore places agent orchestration above deterministic localization, perception, and motion skills; maintains explicit shared context and hierarchical memory; and requires candidate adaptations to pass an offline safety gate. This separation preserves the real-time guarantees of the autonomy stack while enabling voice interaction, semantic reasoning, and long-horizon operational improvement.

\section{Harness Robotic OS Architecture}

As shown in Fig.~\ref{fig:architecture}, Argos is built on Harness Robotic OS, a unified embodied-agent operating architecture rather than a conventional navigation stack. It connects robot hardware abstraction, reusable autonomy skills, agent orchestration, shared context and persistent memory, multimodal reasoning, voice interaction, and property-management workflows through a single closed-loop runtime. The architecture allows sensing, localization, planning, inspection understanding, and operational feedback to be composed as traceable capabilities instead of isolated modules. A governed self-evolution path accumulates execution experience, evaluates failures and task outcomes, produces candidate adaptations, and promotes only validated versions into the deployed skill runtime.

\subsection{Robot Runtime: Vbot Quadruped Platform}

The physical layer uses a Vbot quadruped as the mobile carrier for sensing, computation, communication, and locomotion. The platform is equipped with stereo cameras, a 16-line LiDAR, an IMU, GNSS and 4G/5G connectivity, and an RDK S100P edge computer. The onboard computer hosts perception and agent services, while the robot controller exposes motion commands and state feedback through the HROS robot-runtime interface. This separation prevents higher-level agents from issuing unvalidated low-level actuator commands.

For residential inspection, the platform provides four practical capabilities:

\begin{itemize}
\item Stable traversal over ramps, thresholds, speed bumps, and mildly uneven terrain;
\item Superior maneuverability in narrow passages and temporarily cluttered areas;
\item continuous patrol across residential roads and building-adjacent public spaces;
\item onboard execution on an RDK S100P with six ARM Cortex-A78AE CPU cores and a 128-TOPS Nash BPU.
\end{itemize}

\subsection{Algorithm Layer: SLAM, Perception, and Planning}

The autonomy layer exposes four stateful skills through stable HROS interfaces. Fast-LIO2 consumes LiDAR and IMU streams to build the prior map and estimate the online six-degree-of-freedom pose. Hobot-Stereo produces dense near-field depth, which complements the longer-range geometric structure of LiDAR for low obstacles and partially occluded boundaries. The fused representation is converted into the obstacle format consumed by the planning stack.

PCT-Planner generates a global route through the ordered inspection waypoints on the prior point-cloud map. EGO-Planner receives this route together with the current pose and local obstacle representation, and continuously optimizes a dynamically feasible local trajectory. The global planner therefore preserves mission-level coverage, whereas the local planner handles immediate safety and temporary environmental change. Each capability reports its input timestamp, execution state, confidence or failure code, and output reference to the shared context bus, allowing the cognitive runtime to monitor progress without entering the real-time control loop.

\subsection{Cognitive Agent Runtime: Voice, Memory, and Self-Evolution}

Harness Robotic OS provides a voice-aware interaction plane for both on-site personnel and remote property operators. Streaming automatic speech recognition (ASR) converts spoken instructions into timestamped intent hypotheses, which are grounded with the current robot pose, active mission, visible scene, and access policy before execution. Ambiguous or safety-critical instructions require explicit confirmation. Text-to-speech (TTS) closes the interaction loop by reporting task acceptance, navigation progress, detected hazards, recovery actions, and completion status. Text, voice, images, and enterprise messages therefore enter the same multimodal task interface rather than being handled by isolated applications.

The agent memory subsystem is organized into three complementary levels. Working memory retains short-horizon information such as the current mission, recent dialogue, robot state, local observations, and pending tool calls. Episodic memory stores time-indexed inspection trajectories, decisions, observations, hazard events, failures, and recovery outcomes. Semantic memory maintains stable site knowledge, including map regions, asset identifiers, inspection rules, historical defects, and property-management procedures. Retrieval is conditioned on mission intent, spatial location, scene semantics, and execution state, allowing the agent to reuse relevant experience without placing the complete operational history in every reasoning context.

Self-evolution is implemented as a governed experience-to-update process rather than unconstrained online model modification. After each task, execution traces and human feedback are written to an experience buffer. A reflection and evaluation stage performs outcome scoring, failure attribution, and consistency checks, then proposes candidate updates to memory entries, prompts, tool-selection policies, task graphs, or reusable skills. Candidate versions are evaluated offline against regression cases and safety rules, recorded with provenance, and deployed through versioned rollout only after passing the safety gate. This design enables continuous system improvement while preserving reproducibility, rollback capability, and human operational control.

\subsection{Operational Understanding and Closed-Loop Reporting}

Argos acquires inspection imagery during route traversal and at designated 360\textdegree{} viewpoints. OpenClaw binds each selected image to the current pose, timestamp, waypoint, mission identifier, and applicable inspection policy before invoking Qwen3-VL. The returned description is parsed into a constrained event schema containing the hazard category, severity, evidence, location, and recommended review action. Target categories include blocked fire passages, standing water, exposed wiring, overflowing waste containers, scattered refuse, and abnormal clutter near equipment rooms.

Only schema-valid events are admitted to the operational pipeline. HROS records the original evidence and model response, applies location and regional labels, and generates a structured report. Alerts are then routed to DingTalk or Feishu for human review and work-order processing. This design keeps multimodal inference outside the safety-critical motion loop while preserving the provenance required to audit, correct, or replay each inspection decision.

\section{System Implementation}

\subsection{LiDAR--Inertial Mapping and Relocalization}

\begin{figure}[H]
\centering
\begin{minipage}[t]{0.52\textwidth}
  \vspace{0pt}
  \centering
  \begin{tikzpicture}
    \node[inner sep=0,draw=black!20,line width=0.4pt] (map)
      {\includegraphics[height=4.72cm]{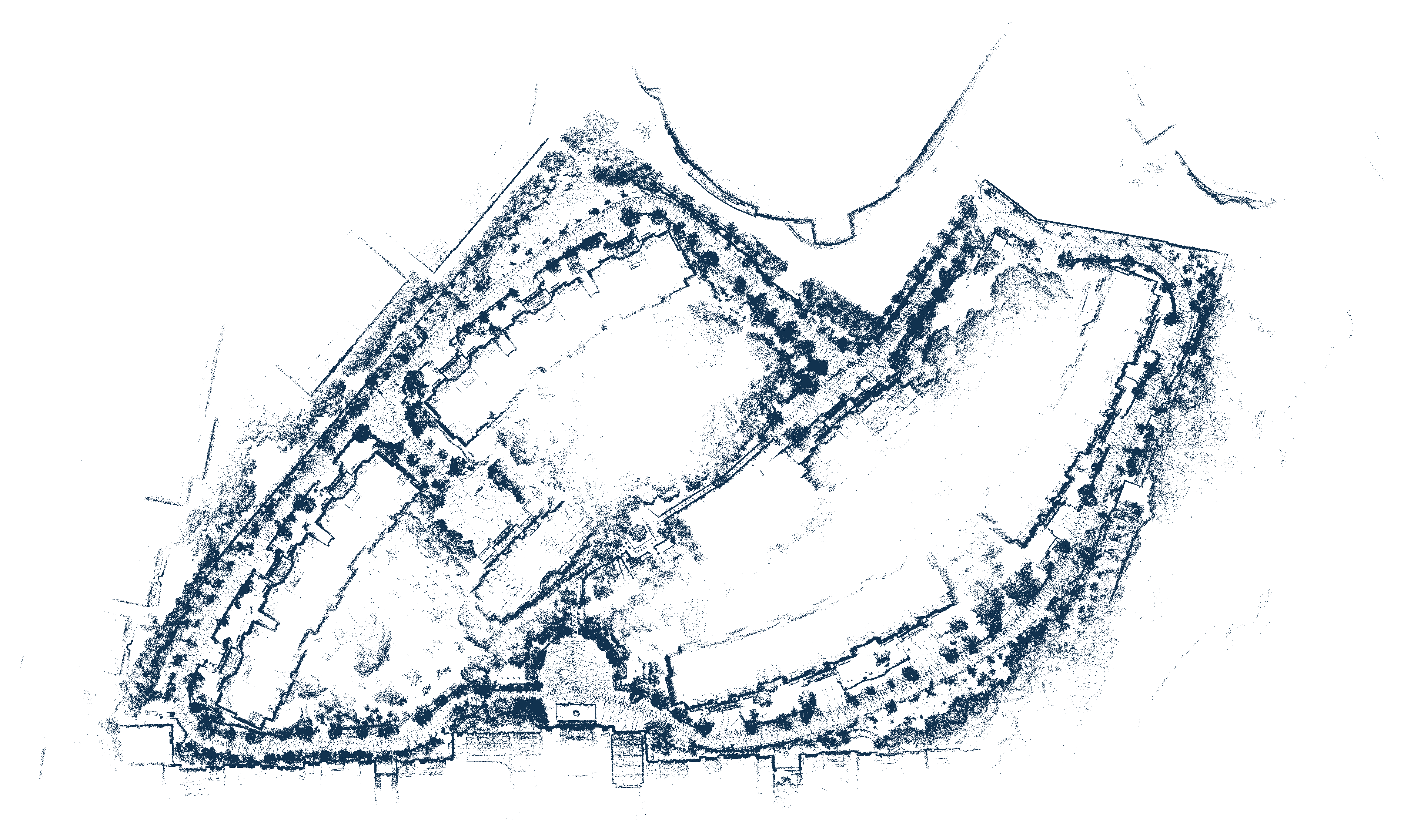}};
    \node[anchor=north west,fill=white,draw=blue!28,rounded corners=1.3pt,
      inner xsep=4pt,inner ysep=2.6pt,font=\sffamily\scriptsize]
      at ([xshift=2.5mm,yshift=-2.5mm]map.north west)
      {\textcolor{blue!65!black}{\textbf{Fast-LIO2}}\enspace Global Point-Cloud Map};
    \draw[-{Latex[length=2mm]},line width=0.75pt,draw=black!72]
      ([xshift=-4mm,yshift=-11mm]map.north east) -- ++(0,7mm)
      node[above,font=\sffamily\scriptsize] {N};
  \end{tikzpicture}
  \par\vspace{0.15em}
  {\small\textbf{(a)} Global LiDAR--inertial map}
\end{minipage}\hfill
\begin{minipage}[t]{0.46\textwidth}
  \vspace{0pt}
  \centering
  \begin{tikzpicture}
    \node[inner sep=0,draw=black!35,line width=0.4pt] (stereo)
      {\includegraphics[height=4.72cm]{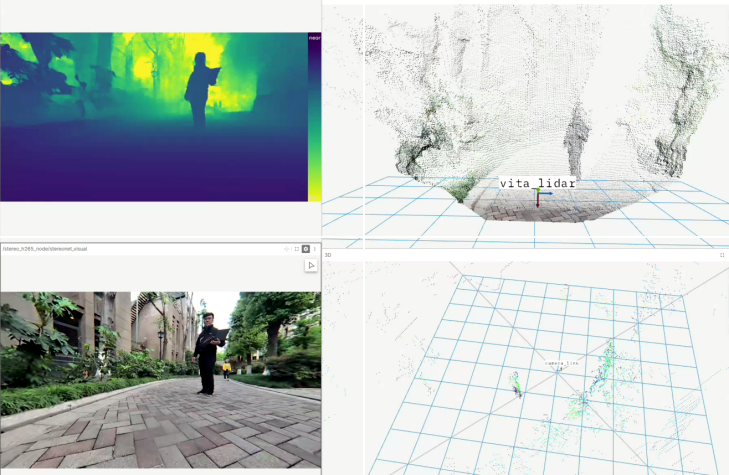}};
    \node[anchor=north west,fill=white,draw=blue!28,rounded corners=1.1pt,
      inner xsep=3pt,inner ysep=2pt,font=\sffamily\scriptsize]
      at ([xshift=1.8mm,yshift=-1.8mm]stereo.north west)
      {\textcolor{blue!65!black}{\textbf{Stereo Depth}}};
    \node[anchor=north west,fill=white,draw=blue!28,rounded corners=1.1pt,
      inner xsep=3pt,inner ysep=2pt,font=\sffamily\scriptsize]
      at ([xshift=1.8mm,yshift=-1.8mm]stereo.north)
      {\textcolor{blue!65!black}{\textbf{Fused Point Cloud}}};
    \node[anchor=north west,fill=white,draw=blue!28,rounded corners=1.1pt,
      inner xsep=3pt,inner ysep=2pt,font=\sffamily\scriptsize]
      at ([xshift=1.8mm,yshift=-1.8mm]stereo.west)
      {\textcolor{blue!65!black}{\textbf{RGB Input}}};
    \node[anchor=north west,fill=white,draw=blue!28,rounded corners=1.1pt,
      inner xsep=3pt,inner ysep=2pt,font=\sffamily\scriptsize]
      at ([xshift=1.8mm,yshift=-1.8mm]stereo.center)
      {\textcolor{blue!65!black}{\textbf{Bird's-Eye View}}};
  \end{tikzpicture}
  \par\vspace{0.15em}
  {\small\textbf{(b)} Local stereo depth and obstacle representation}
\end{minipage}
\caption{Complementary environment representations used by Argos. (a) The global three-dimensional point-cloud map constructed by Fast-LIO2 in the shared mission frame. (b) Hobot-Stereo output showing dense depth (upper left), fused point cloud (upper right), RGB input (lower left), and bird's-eye local geometry (lower right). The global map supports relocalization and mission planning, while dense stereo geometry supplements LiDAR for near-field obstacle perception.}
\label{fig:perception-overview}
\end{figure}
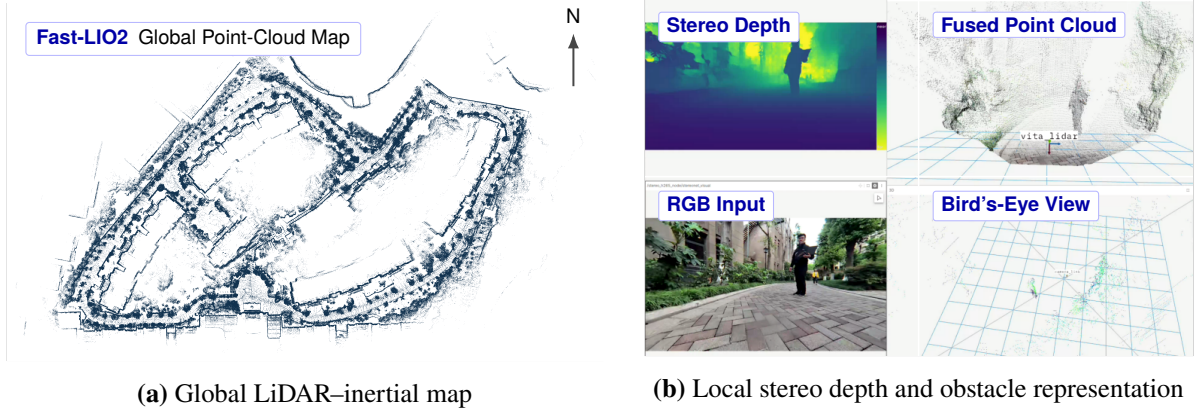

Fast-LIO2 tightly couples LiDAR measurements with inertial observations to estimate motion and incrementally construct a point-cloud map \cite{ref9}. Argos uses the same estimator in three operating modes:

\begin{enumerate}
\item \textbf{Mapping:} during site commissioning, synchronized LiDAR and IMU data are accumulated into the global map shown in Fig.~\ref{fig:perception-overview}(a).
\item \textbf{Online localization:} during routine patrols, the live scan is registered against the prior map to provide the pose used by both planners and the inspection-event logger.
\item \textbf{Relocalization:} after tracking degradation or a restart, the estimator recovers the robot pose in the map frame before autonomous motion resumes.
\end{enumerate}

The shared map frame is essential because a residential patrol repeatedly revisits the same assets under changing illumination and scene appearance. Binding every image, waypoint, hazard event, and report to this frame also allows HROS to retrieve spatially relevant memory and compare observations across missions.

\subsection{Global Inspection Planning}

Property inspection is a coverage-oriented mission rather than a single start-to-goal query. The route must connect policy-defined viewpoints, such as fire-safety facilities, equipment-room entrances, and waste-collection areas, while remaining traversable for the robot. Argos uses PCT-Planner to compute collision-free path segments on the three-dimensional prior map \cite{ref10}. The mission layer orders these segments according to the inspection policy and stores the resulting route as a reusable task template.

At runtime, the global route acts as a reference rather than a direct motion command. Progress is represented by the active segment, current waypoint, completed viewpoints, and remaining inspection actions. This representation allows the agent orchestrator to pause, resume, or re-order non-safety-critical tasks without modifying the local controller.

\subsection{Stereo--LiDAR Obstacle Perception}

Sparse LiDAR sampling can under-represent near-field low obstacles, thin structures, and partial occlusion boundaries. Argos therefore uses Hobot-Stereo to estimate dense depth from synchronized stereo images and fuses the resulting points with the LiDAR obstacle representation. As illustrated in Fig.~\ref{fig:perception-overview}(b), the local depth products complement the global LiDAR geometry in Fig.~\ref{fig:perception-overview}(a): LiDAR provides stable metric structure over a wider range, whereas stereo supplies dense local boundary detail.

Depth samples are transformed into the map frame, filtered by range and confidence, and inserted into the local voxel representation consumed by EGO-Planner. Stereo depth supplements rather than replaces LiDAR; uncertain observations are treated conservatively and expire from the local map when they are not re-observed.

\begin{figure}[H]
\centering
\begin{tikzpicture}
\node[inner sep=0,draw=black!20,line width=0.4pt] (ego) {\includegraphics[width=0.88\textwidth]{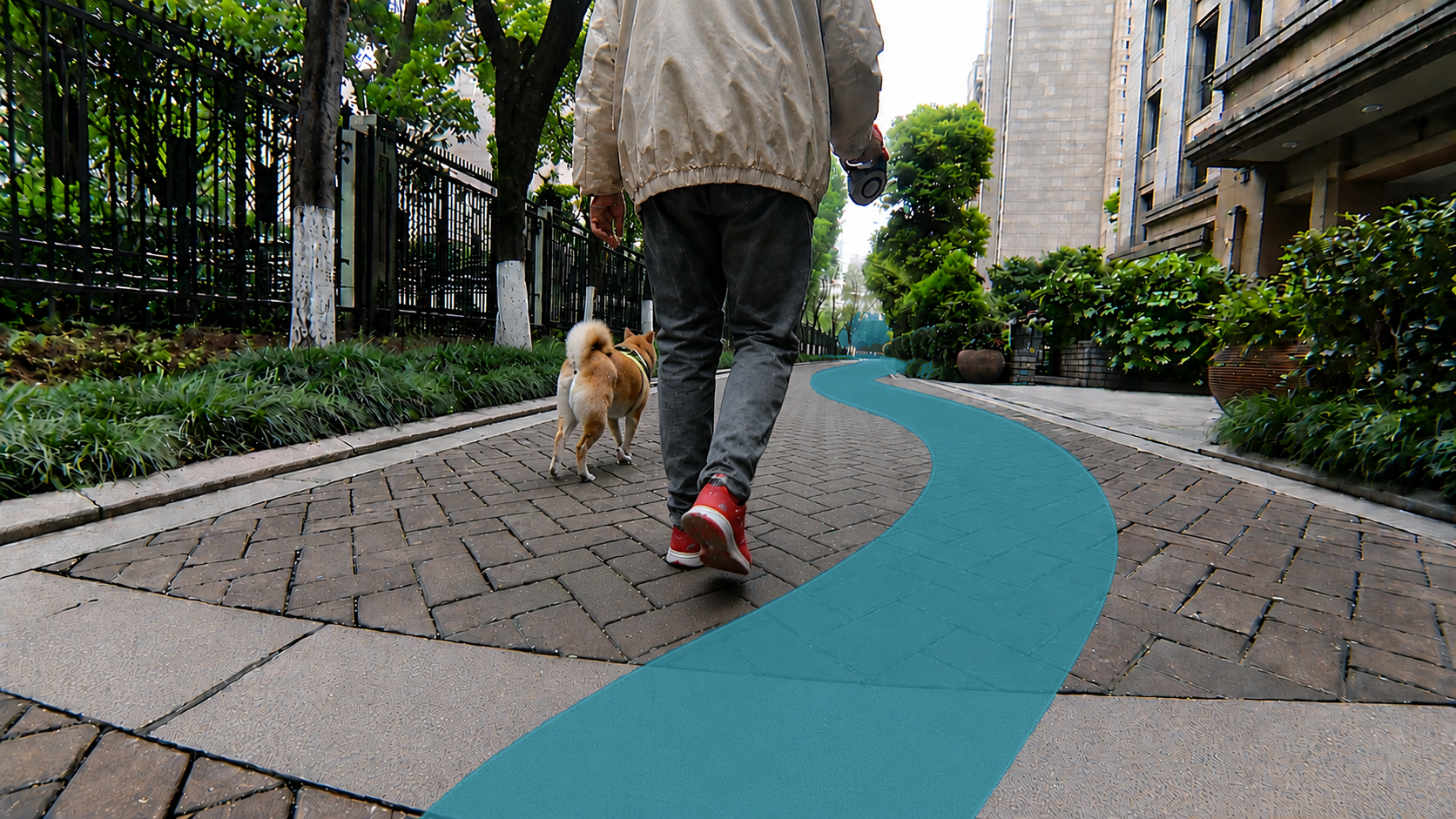}};
\node[anchor=north west,fill=white,draw=black!24,rounded corners=1.5pt,inner sep=3pt,
      font=\sffamily\scriptsize] (egolabel) at ([xshift=2.5mm,yshift=-2.5mm]ego.north west)
      {\textbf{EGO-Planner} \enspace optimized local corridor};
\draw[-{Latex[length=2mm]},draw=blue!65!black,line width=0.9pt]
      (egolabel.south east) to[out=-20,in=115] ([xshift=13mm,yshift=8mm]ego.south);
\end{tikzpicture}
\caption{Obstacle-avoidance corridor generated by EGO-Planner. Blue denotes the locally optimized traversable corridor.}
\label{fig:ego-planner}
\end{figure}

\subsection{Local Motion Generation and Recovery}

EGO-Planner optimizes the local trajectory using the global reference, current robot state, and fused obstacle representation \cite{ref12}. The generated trajectory is converted into commands accepted by the quadruped controller, subject to velocity, clearance, and continuity limits. Figure~\ref{fig:ego-planner} shows a representative locally optimized corridor around an obstacle.

\begin{enumerate}
\item \textbf{Nominal tracking:} follow the current global segment while maintaining obstacle clearance and smooth motion.
\item \textbf{Local replanning:} generate a short detour when pedestrians, parked vehicles, cleaning equipment, or other temporary objects obstruct the reference path.
\item \textbf{Recovery:} stop the platform and report a typed failure when no feasible local trajectory is available, allowing the mission layer to wait, retry, or request operator assistance.
\end{enumerate}

\subsection{Multimodal Hazard Analysis and Enterprise Integration}

The inspection pipeline is designed around an evidence-preserving sequence: observe, interpret, validate, report, and review. OpenClaw selects images associated with a waypoint, attaches pose and mission context, invokes Qwen3-VL, and validates the response against the structured event schema. The deployed taxonomy contains two top-level groups:

\begin{enumerate}
\item \textbf{Safety hazards:} clutter near power-distribution rooms, blocked fire passages, standing water, and exposed wiring;
\item \textbf{Sanitation hazards:} overflowing bins, ground stains, scattered waste or leaves, and abnormal accumulation in public areas.
\end{enumerate}

For each accepted event, the system retains the source image, raw model response, parsed fields, and delivery status. DingTalk and Feishu adapters then distribute the report and associated evidence to the responsible personnel. Human corrections are written back as labeled feedback rather than silently overwriting the original result, enabling subsequent evaluation and memory updates while preserving an auditable record.

\section{Deployment and Evaluation Protocol}

The field evaluation follows the same five-stage lifecycle used in routine deployment. This protocol tests not only individual navigation components but also the transfer of state and evidence across the HROS runtime.

\subsection{Phase 1: Establish 3D Point Cloud Map}

During commissioning, an operator guides the robot through the accessible inspection area while synchronized LiDAR and IMU measurements are recorded. Fast-LIO2 constructs the prior point-cloud map used by localization and route generation. The map is reviewed for gross registration errors before autonomous trials begin.

\begin{figure}[H]
\centering
\begin{tikzpicture}
\node[inner sep=0,draw=black!20,line width=0.4pt] (route) {\includegraphics[width=0.90\textwidth]{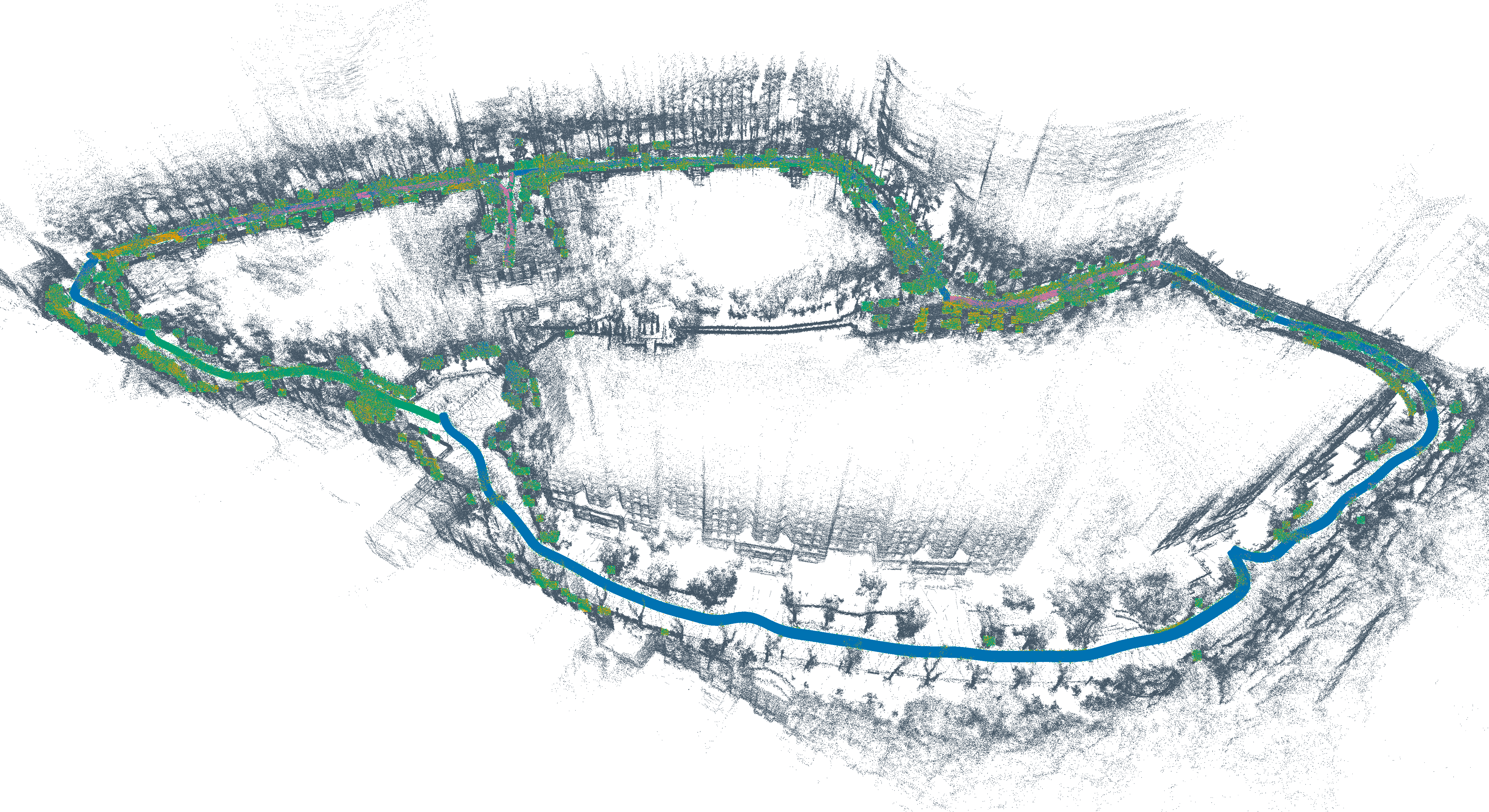}};
\node[anchor=north west,fill=white,draw=black!24,rounded corners=1.5pt,inner sep=3pt,
      font=\sffamily\scriptsize] at ([xshift=2.5mm,yshift=-2.5mm]route.north west)
      {\textbf{PCT-Planner} \enspace global inspection route};
\node[anchor=south east,fill=white,draw=black!20,rounded corners=1.5pt,inner sep=3pt,
      font=\sffamily\scriptsize] at ([xshift=-2.5mm,yshift=2.5mm]route.south east)
      {\tikz\fill[blue!70!black] (0,0) rectangle (2.7mm,1.5mm);\enspace planned route\quad
       \tikz\fill[black!38] (0,0) rectangle (2.7mm,1.5mm);\enspace point-cloud map};
\end{tikzpicture}
\caption{Global inspection route containing eight inspection waypoints. Colorblind-safe route colors are shown against a desaturated point-cloud map.}
\label{fig:global-path}
\end{figure}
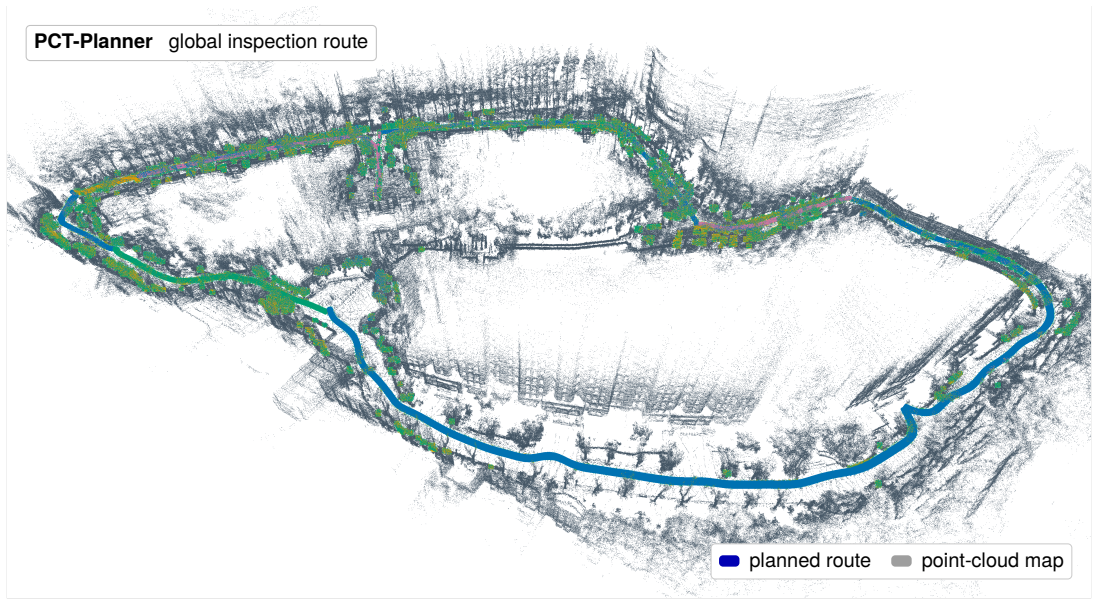

\subsection{Phase 2: Configure the Inspection Mission}

Property personnel define the start, terminal location, inspection regions, and required viewpoints. PCT-Planner connects the ordered viewpoints on the prior map to form the global route. Figure~\ref{fig:global-path} shows the resulting mission with eight inspection waypoints. The route and waypoint actions are stored as a versioned mission template.

\subsection{Phase 3: Daily Autonomous Inspection}

Fast-LIO2 is switched to localization mode and the robot executes the stored mission. EGO-Planner continuously updates the local trajectory using the current pose and fused LiDAR--stereo obstacle representation. The runtime records waypoint state transitions, replanning events, failure codes, and recovery actions.

\subsection{Phase 4: Collect Images and Upload for Analysis}

Images are collected during traversal and at fixed-point 360\textdegree{} inspection actions. OpenClaw selects frames, binds them to the current mission context, and submits standardized requests to Qwen3-VL. Returned results are parsed into the inspection-event schema and checked before operational use.

\subsection{Phase 5: Inspection Result Feedback and Management Closed Loop}

Validated events are associated with timestamps, map locations, region labels, and source evidence. HROS generates a structured inspection report and forwards it through the DingTalk and Feishu adapters for manual review, work-order dispatch, and status tracking. Delivery acknowledgments and operator corrections are retained in the episode record.

\subsection{Controlled Evaluation of Cognitive Runtime}
\label{sec:cognitive-evaluation}

Three controlled experiments are defined to evaluate the HROS capabilities that are not captured by navigation and hazard-recognition measurements alone.

\begin{enumerate}
\item \textbf{Voice interaction:} a fixed set of inspection commands is issued through text and speech under quiet, moderate-noise, and outdoor-noise conditions. The text interface provides the intent reference. We report word error rate, $\mathrm{WER}=(S+D+I)/N$, grounded-intent accuracy, command-completion rate, confirmation accuracy for ambiguous or safety-critical instructions, and the 95th-percentile latency from speech end to acknowledged action. This experiment separates transcription quality from mission-level command understanding.

\item \textbf{Hierarchical agent memory:} repeated inspection cases are evaluated with no persistent memory, working memory only, and the complete working--episodic--semantic memory hierarchy. Queries cover previously observed hazards, spatially grounded asset history, operator corrections, and recovery episodes. We report Recall@$5$ for retrieving the labeled relevant record, spatial--temporal grounding accuracy, 95th-percentile retrieval latency, task-completion rate, context-token reduction, and stale-memory error rate. The principal effect size is the change in task-completion rate relative to the no-memory condition.

\item \textbf{Safety-gated self-evolution:} candidate prompt, memory, tool-policy, and task-graph updates are generated from a training set of execution failures and evaluated on a disjoint regression suite before rollout. Each candidate is compared with its deployed baseline using task-success change, regression rate on previously successful cases, safety-rule violation rate, safety-gate rejection rate, and rollback success rate. An unsafe candidate that reaches the deployed runtime is counted as a gate escape; the required gate-escape rate is zero. This protocol evaluates governed adaptation rather than unconstrained online model modification.
\end{enumerate}

For reproducibility, every reported value should include the number of commands, retrieval queries, missions, candidate updates, and independent repetitions, together with the noise condition or evaluation split. Latency metrics are reported using both the median and 95th percentile, and paired confidence intervals are used for baseline--candidate comparisons.

\section{Experimental Results}

Table~\ref{tab:results} summarizes the aggregate system-level measurements obtained in the residential deployment. Results are organized by operational stage and linked to the subsystem and evaluation role represented by each metric. The cognitive-runtime protocol in Sec.~\ref{sec:cognitive-evaluation} defines the additional measurements required for voice, memory, and governed-adaptation results; numerical values are reported only after the corresponding controlled trials are completed.

\begin{table}[H]
\caption{System-level performance of Argos in the residential inspection deployment. Results are grouped by operational stage to connect component measurements with their role in the end-to-end inspection loop.}
\label{tab:results}
\centering
\scriptsize
\renewcommand{\arraystretch}{1.12}
\begin{tabularx}{\columnwidth}{@{}>{\raggedright\arraybackslash}p{0.19\columnwidth}>{\raggedright\arraybackslash}X>{\centering\arraybackslash\bfseries}p{0.13\columnwidth}>{\raggedright\arraybackslash}p{0.24\columnwidth}@{}}
\toprule
\textbf{Subsystem} & \textbf{Metric} & \textbf{Result} & \textbf{Evaluation role} \\
\midrule
\multicolumn{4}{@{}l}{\textbf{Navigation and Motion}} \\
Mission execution & Waypoint reachability & 100\% & Route completion \\
Fast-LIO2 & Outdoor localization error & $<10$~cm & Spatial grounding \\
EGO-Planner & Obstacle-response latency & $<200$~ms & Reactive avoidance \\
\midrule
\multicolumn{4}{@{}l}{\textbf{Semantic Inspection}} \\
Qwen3-VL & Garbage-overflow detection rate & 95\% & Sanitation hazard \\
Qwen3-VL & Lane-occupancy detection rate & 90\% & Access obstruction \\
Qwen3-VL & Fire-passage obstruction detection rate & 95\% & Fire-safety hazard \\
Qwen3-VL & Standing-water detection rate & 88\% & Environmental hazard \\
Qwen3-VL & Public-facility damage detection rate & 85\% & Asset-condition hazard \\
Inspection inference & Hazard false-positive rate & $<5\%$ & Alert precision \\
Inspection inference & Hazard missed-detection rate & $<5\%$ & Hazard coverage \\
\midrule
\multicolumn{4}{@{}l}{\textbf{Operational Closed Loop}} \\
DingTalk/Feishu adapters & Alarm-delivery success rate & 99\% & Enterprise delivery \\
HROS reporting & Structured-report generation accuracy & 99\% & Report integrity \\
\midrule
\multicolumn{4}{@{}l}{\textbf{Field Operation}} \\
Robot platform & Continuous operating endurance & $>3$~h & Patrol availability \\
End-to-end mission & Full-coverage mission duration & $\leq60$~min & Inspection throughput \\
\bottomrule
\end{tabularx}
\end{table}

\subsection{Navigation and Mission Execution}

All configured waypoints were reached in the reported trials. The outdoor localization error remained below 10~cm, and the measured obstacle-response latency was below 200~ms. Together, these values indicate that the localization and local-planning loop operated at a rate suitable for the tested patrol speed. The endurance exceeded 3~h, while one full-coverage mission required no more than 60~min, leaving operational margin for repeated missions under the evaluated configuration.

\subsection{Hazard Recognition and Operational Delivery}

Detection rates varied with the visual class. Garbage overflow and fire-passage obstruction reached 95\%, lane occupancy reached 90\%, standing water reached 88\%, and public-facility damage reached 85\%. The lower rate for facility damage is consistent with the greater visual diversity of this category. Aggregate false-positive and missed-detection rates were each below 5\%. The communication and reporting stages achieved 99\% success, showing that most accepted events were transferred into the enterprise workflow without loss.

These results establish the feasibility of the deployed Argos pipeline across navigation, semantic inspection, and operational delivery. Future evaluations will extend this analysis to additional properties and longer operating periods, with particular attention to the contribution of memory, voice interaction, and governed adaptation.

\section{Discussion}

\subsection{From Navigation Stack to Embodied-Agent Runtime}

The main system contribution is the explicit separation between deterministic autonomy skills and the cognitive agent runtime. Fast-LIO2, PCT-Planner, Hobot-Stereo, and EGO-Planner remain responsible for time-sensitive state estimation and motion generation. HROS composes these capabilities at the mission level, maintains shared context, and records the provenance of multimodal decisions. Consequently, a reasoning failure cannot directly bypass the motion controller, while navigation events remain accessible to the inspection workflow.

\subsection{From Data Collection to Operational Intelligence}

The value of an inspection robot is determined not only by the area it traverses but also by whether its observations become actionable evidence. Binding images and model outputs to the map frame, mission, and inspection policy allows HROS to produce structured events instead of isolated media files. Enterprise adapters then connect these events to human review and work-order handling. The episode record preserves both the original inference and subsequent correction, creating usable data for later evaluation.

\subsection{Memory and Governed Evolution}

Hierarchical memory provides different retention horizons for active execution, mission history, and stable site knowledge. This structure can support recurring patrols without placing the complete history in every model context. The self-evolution loop is deliberately conservative: feedback may produce candidate changes to memory, prompts, tools, or task graphs, but deployment requires offline evaluation, provenance tracking, and rollback. This governance boundary is necessary for long-running robots whose behavior must remain reproducible and auditable.

\section{Limitations and Future Work}

The current study demonstrates an integrated field system, but several limitations remain:

\begin{enumerate}
\item \textbf{Long-term map maintenance:} parking patterns, construction, vegetation, and seasonal change require incremental mapping, change detection, and multi-session map management.
\item \textbf{Open-world hazard recognition:} broader hazard taxonomies require more diverse labeled data, calibrated confidence, policy constraints, and systematic handling of ambiguous observations.
\item \textbf{Agent evaluation and safety:} the memory and self-evolution mechanisms require dedicated benchmarks for retrieval quality, adaptation benefit, regression risk, and rollback reliability before autonomous updates can be enabled in production.
\item \textbf{Human--robot collaboration:} future deployments should evaluate ASR robustness in outdoor noise, confirmation design for safety-critical commands, operator workload, and integration with access control, broadcasting, alarm, and digital-twin systems.
\end{enumerate}

\section{Conclusion}

This paper presented Harness Robotic OS, a unified runtime for composing robot resources, autonomy skills, multimodal reasoning, memory, human interaction, and enterprise operations, together with Argos, its quadruped inspection realization. The deployed system combines LiDAR--inertial localization, stereo-enhanced obstacle perception, hierarchical global and local planning, contextualized vision-language analysis, and structured operational reporting. Field results demonstrate the feasibility of the navigation and inspection pipeline in a residential property environment, including complete waypoint reachability in the reported trials, sub-10-cm outdoor localization error, sub-200-ms obstacle response, and high delivery reliability for structured events.

Beyond the individual components, HROS defines a boundary between real-time robot control and cognitive adaptation. Shared context and hierarchical memory connect repeated missions, while the safety-gated evolution process provides a path toward improvement without uncontrolled online modification. Future work will extend the evaluation across sites, quantify the contribution of each cognitive component, and validate long-term memory, voice interaction, and governed adaptation under realistic operational noise and change.

\end{document}

%% file: figures/system_architecture.tex
\begin{tikzpicture}[
  font=\sffamily,
  >=Latex,
  shell/.style={rounded corners=4pt, draw=border, fill=surface,
                line width=0.65pt, minimum width=15.2cm},
  sectiontag/.style={rounded corners=8pt, fill=white, draw=border, line width=0.55pt,
                    inner xsep=6pt, inner ysep=2.5pt, text=muted,
                    font=\sffamily\bfseries\fontsize{6.2}{7.2}\selectfont},
  card/.style={rounded corners=3pt, draw=border, fill=white, line width=0.65pt,
               text width=3.70cm, minimum height=0.90cm, align=center, inner sep=3.5pt,
               font=\sffamily\fontsize{6.9}{8.3}\selectfont},
  compact/.style={rounded corners=3pt, draw=border, fill=white, line width=0.65pt,
                  text width=2.88cm, minimum height=0.86cm, align=center, inner sep=3pt,
                  font=\sffamily\fontsize{6.55}{7.85}\selectfont},
  memorycard/.style={compact, draw=violet!55, fill=violet!3},
  evolvecard/.style={rounded corners=2.5pt, draw=violet!30, fill=white,
                     text width=2.86cm, minimum height=0.68cm, align=center, inner sep=2.5pt,
                     font=\sffamily\fontsize{6.15}{7.25}\selectfont},
  data/.style={-{Latex[length=2.05mm,width=1.4mm]}, draw=navy, line width=0.78pt},
  binding/.style={{Latex[length=1.8mm]}-{Latex[length=1.8mm]}, draw=cyan,
                  line width=0.76pt},
  evolve/.style={-{Latex[length=2.0mm,width=1.35mm]}, draw=violet,
                 line width=0.82pt, densely dashed},
  guard/.style={-{Latex[length=2.0mm,width=1.35mm]}, draw=mint,
                line width=0.82pt}
]
\definecolor{navy}{HTML}{102A43}
\definecolor{indigo}{HTML}{4057D6}
\definecolor{cyan}{HTML}{0797B8}
\definecolor{violet}{HTML}{7656D6}
\definecolor{mint}{HTML}{19A99B}
\definecolor{surface}{HTML}{F7F9FC}
\definecolor{border}{HTML}{CBD5E1}
\definecolor{muted}{HTML}{53657A}

\node[rounded corners=4pt,draw=border,fill=white,minimum width=15.2cm,minimum height=0.74cm,
      line width=0.7pt] at (7.60,9.05) {};
\node[circle,fill=indigo,text=white,minimum size=0.48cm,inner sep=0pt,
      font=\fontsize{8.5}{9}\selectfont] at (0.70,9.05) {\faRobot};
\node[anchor=west,text=navy,font=\sffamily\bfseries\fontsize{10.3}{11.6}\selectfont]
      at (1.08,9.05) {HARNESS ROBOTIC OS};
\node[anchor=east,text=muted,font=\sffamily\fontsize{6.8}{8}\selectfont]
      at (14.96,9.05) {voice-aware $\cdot$ memory-augmented $\cdot$ self-evolving embodied agents};

\node[shell,minimum height=1.30cm] at (7.60,7.72) {};
\node[sectiontag] at (1.70,8.39) {INTERACTION AND OPERATIONS};
\node[card] (voice) at (2.70,7.72)
      {{\color{cyan}\faMicrophone}\, {\color{cyan}\faVolumeUp}\quad\textbf{Voice and Multimodal I/O}\\
       streaming ASR $\cdot$ TTS $\cdot$ text/image};
\node[card] (mission) at (7.60,7.72)
      {{\color{indigo}\faTasks}\quad\textbf{Inspection Mission Console}\\
       policy $\cdot$ waypoint task $\cdot$ human confirmation};
\node[card] (enterprise) at (12.50,7.72)
      {{\color{mint}\faComments}\quad\textbf{Enterprise Closed Loop}\\
       report $\cdot$ alert $\cdot$ DingTalk/Feishu $\cdot$ voice feedback};

\node[rounded corners=5pt,draw=indigo!72,fill=indigo!3,minimum width=15.2cm,
      minimum height=3.45cm,line width=0.95pt] at (7.60,5.22) {};
\node[rounded corners=7pt,fill=indigo,text=white,inner xsep=8pt,inner ysep=3pt,
      font=\sffamily\bfseries\fontsize{7.4}{8.7}\selectfont] at (3.05,6.78)
      {\faLayerGroup\quad COGNITIVE AGENT RUNTIME};
\node[anchor=east,text=muted,font=\sffamily\fontsize{6.0}{7.2}\selectfont]
      at (14.88,6.78) {shared context, persistent memory, governed adaptation};

\node[compact] (orchestrator) at (2.15,5.74)
      {{\color{indigo}\faUserCog}\quad\textbf{Agent Orchestrator}\\OpenClaw $\cdot$ intent routing $\cdot$ task graph};
\node[memorycard] (memory) at (5.78,5.74)
      {{\color{violet}\faMemory}\quad\textbf{Hierarchical Agent Memory}\\working $\cdot$ episodic $\cdot$ semantic};
\node[compact] (reasoning) at (9.41,5.74)
      {{\color{violet}\faBrain}\quad\textbf{Multimodal Reasoning}\\Qwen3-VL\\retrieval $\cdot$ reflection};
\node[compact] (skills) at (13.04,5.74)
      {{\color{cyan}\faPuzzlePiece}\quad\textbf{Skill and Tool Runtime}\\compose $\cdot$ call $\cdot$ monitor $\cdot$ recover};
\draw[data] (orchestrator.east) -- (memory.west);
\draw[data] (memory.east) -- (reasoning.west);
\draw[data] (reasoning.east) -- (skills.west);

\node[rounded corners=3pt,draw=violet!24,fill=violet!5,minimum width=14.45cm,
      minimum height=1.18cm,line width=0.55pt] at (7.60,4.05) {};
\node[sectiontag,draw=violet!24,text=violet,anchor=west]
      at (0.48,4.64) {\faRedo\quad SELF-EVOLUTION LOOP};
\node[evolvecard] (experience) at (2.15,3.95)
      {{\color{violet}\faHistory}\quad\textbf{Experience Buffer}\\episodes $\cdot$ traces $\cdot$ failures};
\node[evolvecard] (evaluation) at (5.78,3.95)
      {{\color{violet}\faFlask}\quad\textbf{Reflection and Evaluation}\\success metrics $\cdot$ error attribution};
\node[evolvecard] (update) at (9.41,3.95)
      {{\color{violet}\faCodeBranch}\quad\textbf{Candidate Evolution}\\memory $\cdot$ prompt $\cdot$ skill/task graph};
\node[evolvecard,draw=mint!55,fill=mint!4] (safety) at (13.04,3.95)
      {{\color{mint}\faCheckCircle}\quad\textbf{Safety Gate}\\offline validation $\cdot$ versioned rollout};
\draw[evolve] (experience.east) -- (evaluation.west);
\draw[evolve] (evaluation.east) -- (update.west);
\draw[guard] (update.east) -- (safety.west);
\draw[evolve] (orchestrator.south) -- (experience.north);
\draw[guard] (safety.north) to[out=90,in=-45] (skills.south east);

\node[shell,minimum height=1.20cm] at (7.60,2.65) {};
\node[sectiontag] at (2.15,3.28) {EMBODIED AUTONOMY SKILLS};
\node[compact] (planning) at (2.15,2.65)
      {{\color{cyan}\faRoute}\quad\textbf{Mission Planning}\\PCT-Planner $\cdot$ waypoint sequencing};
\node[compact] (mapping) at (5.78,2.65)
      {{\color{cyan}\faMap}\quad\textbf{State Estimation}\\Fast-LIO2\\mapping $\cdot$ localization};
\node[compact] (perception) at (9.41,2.65)
      {{\color{cyan}\faEye}\quad\textbf{Local Perception}\\Hobot-Stereo $\cdot$ obstacle fusion};
\node[compact] (motion) at (13.04,2.65)
      {{\color{cyan}\faRobot}\quad\textbf{Motion Intelligence}\\EGO-Planner $\cdot$ control $\cdot$ recovery};

\node[shell,minimum height=1.12cm] at (7.60,1.10) {};
\node[sectiontag] at (1.72,1.69) {ROBOT RUNTIME};
\node[compact] (edge) at (2.15,1.10)
      {{\color{mint}\faMicrochip}\quad\textbf{Edge Compute}\\RDK S100P $\cdot$ ASR/TTS/model services};
\node[compact] (sensors) at (5.78,1.10)
      {{\color{mint}\faCamera}\quad\textbf{Multimodal Sensing}\\stereo $\cdot$ LiDAR $\cdot$ IMU $\cdot$ microphone};
\node[compact] (network) at (9.41,1.10)
      {{\color{mint}\faNetworkWired}\quad\textbf{Connectivity and I/O}\\GNSS $\cdot$ 4G/5G $\cdot$ enterprise APIs};
\node[compact] (actuation) at (13.04,1.10)
      {{\color{mint}\faCogs}\quad\textbf{Quadruped Actuation}\\locomotion $\cdot$ posture\\payload};

\draw[binding] (voice.south) -- (orchestrator.north);
\draw[binding] (mission.south) -- (reasoning.north);
\draw[binding] (enterprise.south) -- (skills.north);
\draw[binding] (planning.north west) -- (0.30,3.34)
               -- (0.30,5.74) -- (orchestrator.west);
\draw[binding] (mapping.north west) -- (4.02,3.34)
               -- (4.02,5.10) -- (memory.south west);
\draw[binding] (perception.north west) -- (7.60,3.34)
               -- (7.60,5.10) -- (reasoning.south west);
\draw[binding] (motion.north west) -- (11.22,3.34)
               -- (11.22,5.10) -- (skills.south west);
\draw[data] (edge.north) -- (planning.south);
\draw[data] (sensors.north) -- (mapping.south);
\draw[data] (network.north) -- (perception.south);
\draw[data] (actuation.north) -- (motion.south);

\node[anchor=north east,text=muted,font=\sffamily\fontsize{5.75}{6.8}\selectfont]
      at (15.12,0.43)
      {\tikz\draw[data] (0,0)--(0.52,0); runtime flow\quad
       \tikz\draw[binding] (0,0)--(0.52,0); agent binding\quad
       \tikz\draw[evolve] (0,0)--(0.52,0); evolution\quad
       \tikz\draw[guard] (0,0)--(0.52,0); safety gate};
\end{tikzpicture}